\documentclass[10pt,twocolumn,letterpaper]{article}

\usepackage{cvpr}
\usepackage{times}
\usepackage{epsfig}
\usepackage{graphicx}
\usepackage{amsmath}
\usepackage{amssymb}
\usepackage{subfigure}
\usepackage{array}
\usepackage{url}
\usepackage{authblk}

\usepackage[breaklinks=true,bookmarks=false]{hyperref}

\cvprfinalcopy 

\def\cvprPaperID{****} 

\ifcvprfinal\fi
\begin{document}
\newcolumntype{L}[1]{>{\raggedright\arraybackslash}p{#1}}
\newcolumntype{C}[1]{>{\centering\arraybackslash}p{#1}}
\newcolumntype{R}[1]{>{\raggedleft\arraybackslash}p{#1}}
\title{Domain-Symmetric Networks for Adversarial Domain Adaptation}
\author[1,2] {Yabin Zhang}
\author[1] {Hui Tang}
\author[1] {Kui Jia} 
\author[1] {Mingkui Tan}
\affil[1] {South China University of Technology}
\affil[2] {DAMO Academy, Alibaba Group}
\affil[ ] {\tt\small \{zhang.yabin,eehuitang\}@mail.scut.edu.cn, \{kuijia,mingkuitan\}@scut.edu.cn}



\maketitle
\thispagestyle{empty}

\begin{abstract}
Unsupervised domain adaptation aims to learn a model of classifier for unlabeled samples on the target domain, given training data of labeled samples on the source domain. Impressive progress is made recently by learning invariant features via domain-adversarial training of deep networks. In spite of the recent progress, domain adaptation is still limited in achieving the invariance of feature distributions at a finer category level. To this end, we propose in this paper a new domain adaptation method called Domain-Symmetric Networks (SymNets). The proposed SymNet is  based on a symmetric design of source and target task classifiers, based on which we also construct an additional classifier that shares with them its layer neurons. To train the SymNet, we propose a novel adversarial learning objective whose key design is based on a two-level domain confusion scheme, where the category-level confusion loss improves over the domain-level one by driving the learning of intermediate network features to be invariant at the corresponding categories of the two domains. Both domain discrimination and domain confusion are implemented based on the constructed additional classifier. Since target samples are unlabeled, we also propose a scheme of cross-domain training to help learn the target classifier. Careful ablation studies show the efficacy of our proposed method. In particular, based on commonly used base networks, our SymNets achieve the new state of the art on three benchmark domain adaptation datasets.
\end{abstract}

\section{Introduction} \label{Sec:intro}

Deep learning methods have achieved great success in various machine learning tasks. A common pre-requisite for such success is the availability of massive amounts of annotated training data. For many other tasks, however, these training data are either difficult to collect, or annotating them costs prohibitively. Therefore, to address the scarcity of annotated data on some \emph{target} tasks/domains, there is a strong motivation to leverage the massively available annotated data on related \emph{source} ones via a manner of transfer learning or domain adaptation \cite{transfer_survey}. Unfortunately, this attractive learning paradigm suffers from the problem of domain shift \cite{domain_shift}, which stands as a major obstacle for adapting the learned models on source domains to be useful for target ones.

Domain adaptation aims to obtain models that have smaller risks on target data. Theoretical analysis \cite{da_theory} suggests that such a target risk can be minimized by bounding the risk of a model on the source data and the discrepancy between distributions of the two domains, which inspires many of existing methods \cite{ddc,wmmd,dan,cmd,deep_coral,reverse_grad,dann,domain_confusion,adda,attention_alignment,SimNet}. Among existing methods, those based on domain-adversarial training of deep networks \cite{reverse_grad,dann} achieve the current state of the art on many benchmark domain adaptation datasets \cite{office_31,ImageCLEFDA,office_home}. Inspired by generative adversarial networks \cite{gan}, domain-adversarial training typically plays a minimax game to learn a domain discriminator, which aims to distinguish features of source samples from those of target samples, and a feature extractor, which aims to learn domain-invariant feature representations in order to confuse the domain discriminator. Domain alignment is expected when the minimax optimization reaches an equilibrium.

In spite of the remarkable empirical results achieved by domain-adversarial training methods, they still suffer from a major limitation: even though the feature extractor is well trained to give domain-invariant features of both the source and target samples, the corresponding model/classifier is trained on the source samples and cannot perfectly generalize to the target ones, i.e., the joint distributions of feature and category are not well aligned across data domains. Some of existing methods have paid attention to this issue. For example, in \cite{asy_tri,collaborative,semantic_align}, pseudo labels are assigned to target samples, on which the category-level alignment is promoted.  In \cite{cada,mada}, multiplicative interactions between feature representations and category predictions are exploited as high-order features to help adversarial training.  

These existing methods have to some extent alleviated the above issue. To push further along this line, we propose in this paper a novel design of Domain-Symmetric Networks (SymNets) to facilitate, via adversarial training, the alignment of joint distributions of feature and category across data domains. Similar to \cite{rtn}, our proposed SymNet contains an explicit task classifier for the target domain. Different from \cite{rtn}, we also construct an additional classifier that shares its neurons with those of the source and target classifiers (cf. Section \ref{Sec:symmetric_classifiers} for how the three classifiers are constructed). In this work, we propose a novel adversarial learning method to train the thus constructed SymNet, which includes category-level and domain-level confusion losses and can thus enhance domain-invariant feature learning towards the category level. To make the target classifier more symmetric with the source one in terms of predicting task categories, we also propose a cross-domain training scheme to help training of the target classifier. Careful ablation studies show the efficacy of key designs of our proposed SymNet.

We summarize our main contributions as follows.
\begin{itemize}
\item We propose in this paper a novel method termed SymNet for adversarial domain adaptation. Our proposed SymNet is based on a symmetric design of source and target task classifiers, based on which we also construct an additional classifier that shares with them its layer neurons. Both domain discrimination and domain confusion are implemented based on the constructed additional classifier.  
    
\item To train the SymNet, we propose a novel adversarial learning method based on two-level domain confusion losses, where the category-level confusion loss improves over the domain-level one by driving the learning of intermediate network features to be invariant at the corresponding categories of the two domains. Since target samples are unlabeled, we also propose a scheme of cross-domain training to help learn the target classifier.

\item We conduct careful ablation studies to investigate the efficacy of key designs of our proposed SymNet. These studies empirically corroborate our designs. In particular, based on commonly used base networks, our proposed SymNets achieve the new state of the art on benchmark domain adaptation datasets of Office-31 \cite{office_31}, ImageCLEF-DA \cite{ImageCLEFDA}, and Office-Home \cite{office_home}.

\end{itemize}

\begin{figure*}
\begin{center}
\includegraphics[width=0.65\linewidth] {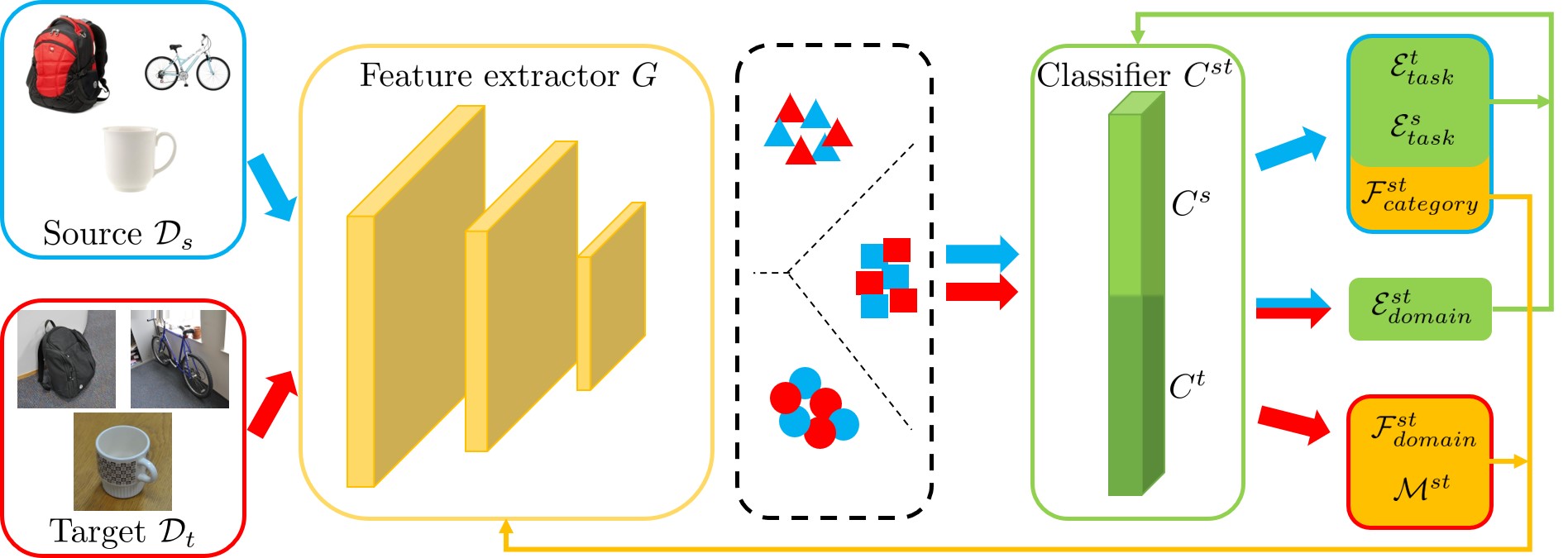}
\end{center}
   \caption{The architecture of our proposed SymNet, which includes a feature extractor $\mathnormal{G}$ and three classifiers of $\mathnormal{C}^s, \mathnormal{C}^t$ and $\mathnormal{C}^{st}$. Note that the classifier $\mathnormal{C}^{st}$ shares its layer neurons with $\mathnormal{C}^s$ and $\mathnormal{C}^t$. The red and blue colors indicate the target data and source data, and the losses generated by them, respectively. The yellow and green colors represent the feature extractor and classifiers, and the losses applied to them, respectively. The middle dashed rectangle presents a toy example of the features of the SymNet that are invariant at the corresponding categories of two domains. Please refer to the main text for how the two-level domain confusion training objectives are defined.}
\label{fig:framework}
\end{figure*}

\section{Related Works}
In this section, we briefly review recent domain adaptation methods, in particular those aiming to align the joint distributions of feature and category across two data domains \cite{asy_tri, collaborative, semantic_align, rtn, mcd, cada, mada}.

Existing domain adaptation methods \cite{ddc, dan,cmd, deep_coral, reverse_grad,dann, domain_confusion, adda, rtn, asy_tri, collaborative, semantic_align, cada, mada} typically learn domain invariant features to minimize the domain discrepancy.
Some of the existing methods \cite{ddc, dan, cmd, deep_coral, reverse_grad, dann, adda} neglect the alignment between the corresponding categories of the two domains. In contrast, to align the joint distributions of feature and category across two data domains, Saito \emph{et al.} \cite{asy_tri} proposes to asymmetrically use three task classifiers, where two task classifiers are utilized to label the unlabeled target samples according to their prediction consistency and the confidence, and another task classifier is trained by these target samples with pseudo labels. However, the trueness of pseudo labels is doubtful and false labels have a profoundly negative impact on the performance. To improve the reliability of pseudo labels for target samples, Zhang \emph{et al.} \cite{collaborative} reweights the target samples by the degree of confusion between domains, specifically those target samples which well confuse the domain discriminator in domain labels, are thus assigned by higher weights. Xie \emph{et al.} \cite{semantic_align} aligns the centroid of each category between the two domains, instead of treating the pseudo labels as true ones directly. Long \emph{et al.} \cite{rtn} uses a residual function to model the shift between the learned task classifiers of the two domains, which can be useful in the adaptation tasks of small domain discrepancy but inadequate to tackle the large domain discrepancy. In \cite{cada,mada}, multiplicative interactions between feature representations and category predictions are exploited as high-order features to help adversarial training. By taking the category decision boundaries into account, Saito \emph{et al.} \cite{mcd} proposes to detect the target samples near the category decision boundaries by maximizing the discrepancy between the outputs of two separate task classifiers and learn a feature extractor to generate features near the source support for these target samples to minimize the discrepancy.

To further promote the alignment of joint distributions of feature and category across data domains, our SymNets contain an explicit task classifier for the target domain and an additional classifier to enable domain discrimination and domain confusion, and have two-level domain confusion losses, where the category-level confusion loss improves over the domain-level one by driving the learning of intermediate network features to be invariant at the corresponding categories of the two domains.

\section{The Proposed Domain-Symmetric Networks} \label{Sec:SymNet}


In unsupervised domain adaptation, we are given a source domain $\mathcal{D}_s = \{(\mathrm{\mathbf{x}}_i^s, y_i^s)\}_{i=1}^{n_s}$ of $n_s$ labeled samples and a target domain $\mathcal{D}_t = \{(\mathrm{\mathbf{x}}_j^t)\}_{j=1}^{n_t}$ of $n_t$ unlabeled samples. The i.i.d. assumption is violated as the source domain $\mathcal{D}_s$ and target domain $\mathcal{D}_t$ are assumed to be different. The goal of unsupervised domain adaptation is to learn a feature extractor $\mathnormal{G}$ and a classifier $\mathnormal{C}$ such that the expected target risk ${\mathbb{E}}_{(\mathbf{x}^t, y^t)\sim {\cal{D}}_t}[{\cal{L}}(C(G(\mathbf{x}^t)), y^t)]$ can be minimized for a certain loss $\cal{L}$.

Theoretical analysis  \cite{da_theory} suggests that the target risk can be minimized by bounding the source risk and the discrepancy between two domains. Inspired by GANs \cite{gan}, domain-adversarial training \cite{reverse_grad,dann} is explored to achieve the later objective. As summarized in \cite{adda}, there are three ways to implement the domain-adversarial training losses: minimax \cite{reverse_grad,dann}, confusion \cite{domain_confusion}, and GAN \cite{adda}. We introduce the domain confusion loss \cite{domain_confusion} that is most related to our method.

Given a deep neural network that is composed of convolutional and fully-connected (FC) layers, the domain confusion method uses the lower convolutional layers as the feature extractor $\mathnormal{G}$ and upper FC layers as the task classifier $\mathnormal{C}$. The domain discriminator $\mathnormal{D}$, which is in parallel with $\mathnormal{C}$, is added on top of $\mathnormal{G}$ to distinguish features of samples from the two domains.
Source risk minimization is achieved based on a standard supervised classification objective:
\begin{equation} \label{Loss_conf_cla}
\min_{\mathnormal{C,G}} {\cal{E}}_{task} = \frac{1}{n_s} \sum_{i=1}^{n_s} {\cal{L}}^s\left(C(G(\mathbf{x}_i^s)), y_i^s\right),
\end{equation}
where $\mathcal{L}^s$ is typically a cross-entropy loss. Due to the existence of domain discrepancy, there is a large drop in performance when directly applying the model trained by (\ref{Loss_conf_cla}) to the target data. Given feature representations of different domains extracted by $\mathnormal{G}$, we can learn a domain discriminator $\mathnormal{D}$ using the following objective:
\begin{align} \label{Loss_conf_dis}
\notag \min_{\mathnormal{D}} {\cal{E}}_{domain} =  & - \frac{1}{n_s}\sum_{i=1}^{n_s} \mathrm{log}(1 - D(G(\mathbf{x}_i^s)) \\ & - \frac{1}{n_t}\sum_{j=1}^{n_t} \mathrm{log}(D(G(\mathbf{x}_j^t)).
\end{align}
Given a $\mathnormal{D}$, the domain confusion loss aims to learn $\mathnormal{G}$ to maximally ``confuse'' the two domains, by computing the cross entropy between the domain predictions and a uniform distribution over domain labels: 
\begin{align} \label{Loss_conf_ali}
\notag \min_{\mathnormal{G}} {\cal{F}}_{domain} = & \frac{1}{2}{\cal{E}}_{domain} - \frac{1}{2n_s}\sum_{i=1}^{n_s} \mathrm{log}(D(G(\mathbf{x}_i^s)) \\ & - \frac{1}{2n_t}\sum_{j=1}^{n_t} \mathrm{log}(1 - D(G(\mathbf{x}_j^t)).
\end{align}
Domain alignment is achieved by learning a domain-invariant $\mathnormal{G}$ based on the following adversarial objective of domain confusion:
\begin{align} \label{Loss_confusion}
\notag \min_{\mathnormal{G,C}} & {\cal{E}}_{task}(G, C) + \lambda {\cal{F}}_{domain}(G, D) \\
\min_{\mathnormal{D}} & {\cal{E}}_{domain} (G, D),
\end{align}
where $\lambda$ is a trade-off parameter.

\subsection{A Symmetric Design of Source and Target Task Classifiers} \label{Sec:categorical_alignment} 
\label{Sec:symmetric_classifiers}

As discussed in Section \ref{Sec:intro}, although impressive results are obtained by existing methods of domain-adversarial training, they still suffer from the fundamental challenge of unsupervised domain adaptation, i.e., the joint distributions of feature and category cannot be well aligned across data domains. To address this challenge, we propose in this paper a novel domain-symmetric network (SymNet), with the corresponding domain-adversarial training method. We first present architectural design of our proposed SymNet as follows (cf. Figure \ref{fig:framework} for an illustration).

The design of a SymNet starts with two parallel task classifiers $\mathnormal{C}^s$ and $\mathnormal{C}^t$. Assume each of the two classifiers is based on a single FC layer (with a subsequent softmax operation). $\mathnormal{C}^s$ and $\mathnormal{C}^t$ respectively contain $K^s$ and $K^t$ neurons corresponding to the numbers of categories on the source and target domains. In unsupervised domain adaptation, we have $K^s = K^t = K$. For an input $\mathbf{x}$ of the SymNet, we respectively denote as $\mathbf{v}^s(\mathbf{x}) \in \mathbb{R}^K$ and $\mathbf{v}^t(\mathbf{x}) \in \mathbb{R}^K$ the output vectors of $\mathnormal{C}^s$ and $\mathnormal{C}^t$ before the softmax operation, and $\mathbf{p}^s(\mathbf{x}) \in [0, 1]^K$ and $\mathbf{p}^t(\mathbf{x}) \in [0, 1]^K$ after the softmax operation. Except for $\mathnormal{C}^s$ and $\mathnormal{C}^t$, our SymNet also has a classifier $\mathnormal{C}^{st}$ whose design is as follows. Given $\mathbf{v}^s(\mathbf{x})$ and $\mathbf{v}^t(\mathbf{x})$ for an input $\mathbf{x}$, we first concatenate them to form $[\mathbf{v}^s(\mathbf{x}); \mathbf{v}^t(\mathbf{x})] \in \mathbb{R}^{2K}$, and we then apply the softmax operation to the concatenated vector to have a probability vector $\mathbf{p}^{st}(\mathbf{x}) \in [0, 1]^{2K}$. We thus have $C^{st}(G(\mathbf{x})) = \mathbf{p}^{st}(\mathbf{x})$. For ease of subsequent notations, we also write $p^s_k(\mathbf{x})$ (\emph{resp.} $p^t_k(\mathbf{x})$ or $p^{st}_k(\mathbf{x})$), $k \in \{1, \dots, K\}$, for the $k^{th}$ element of the category probability vector $\mathbf{p}^s(\mathbf{x})$ (\emph{resp.} $\mathbf{p}^t(\mathbf{x})$ or $\mathbf{p}^{st}(\mathbf{x})$) predicted by $C^s(G(\mathbf{x}))$ (\emph{resp.} $C^t(G(\mathbf{x}))$ or $C^{st}(G(\mathbf{x}))$).

Note that there exists no an explicit domain discriminator in our design of SymNet. Both the domain discrimination and domain confusion is achieved by applying appropriate losses to the classifier $C^{st}$, which we will present shortly. We first present in the following how to train $C^s$ and $C^t$.

\vspace{0.01cm}
\noindent\textbf{Learning of Source Task Classifier}

The task classifier $\mathnormal{C}^s$ is trained using the following cross-entropy loss over the labeled source samples:
\begin{equation} \label{Loss_task_sc}
\min_{\mathnormal{C^s}} {\cal{E}}_{task}^s (G, C^s) = - \frac{1}{n_s} \sum_{i=1}^{n_s} \mathrm{log}(p^s_{y_i^s}(\mathbf{x}_i^s)).
\end{equation}

\noindent\textbf{Cross-Domain Learning of Target Task Classifier}

Since target samples are unlabeled, there exist no direct supervision signals to learn a task classifier $\mathnormal{C}^t$. Our idea is to leverage the labeled source samples, and use the following cross-entropy loss to train $\mathnormal{C}^t$:
\begin{equation} \label{Loss_task_tc}
\min_{\mathnormal{C^t}} {\cal{E}}_{task}^t (G, C^t) = - \frac{1}{n_s} \sum_{i=1}^{n_s} \mathrm{log}(p^t_{y_i^s}(\mathbf{x}_i^s)).
\end{equation}
At a first glance, it seems that (\ref{Loss_task_tc}) learns $\mathnormal{C}^t$ that is a duplicate of $\mathnormal{C}^s$. However, a domain discrimination training via $\mathnormal{C}^{st}$ will make them distinguishable. In fact, the use of (\ref{Loss_task_tc}) is essential to establish a neuron-wise correspondence between $\mathnormal{C}^s$ and $\mathnormal{C}^t$, which provides the basis to achieve category-level domain confusion presented in Section \ref{Sec:two_level_confusion}. The use of labeled source samples in (\ref{Loss_task_tc}) also makes the learned $\mathnormal{C}^t$ more discriminative among task categories. We present ablation studies in Section \ref{Sec:analysis} that confirm the efficacy of our way of learning the target task classifier $\mathnormal{C}^t$.


\begin{figure}
\begin{center}
\includegraphics[width=0.7\linewidth] {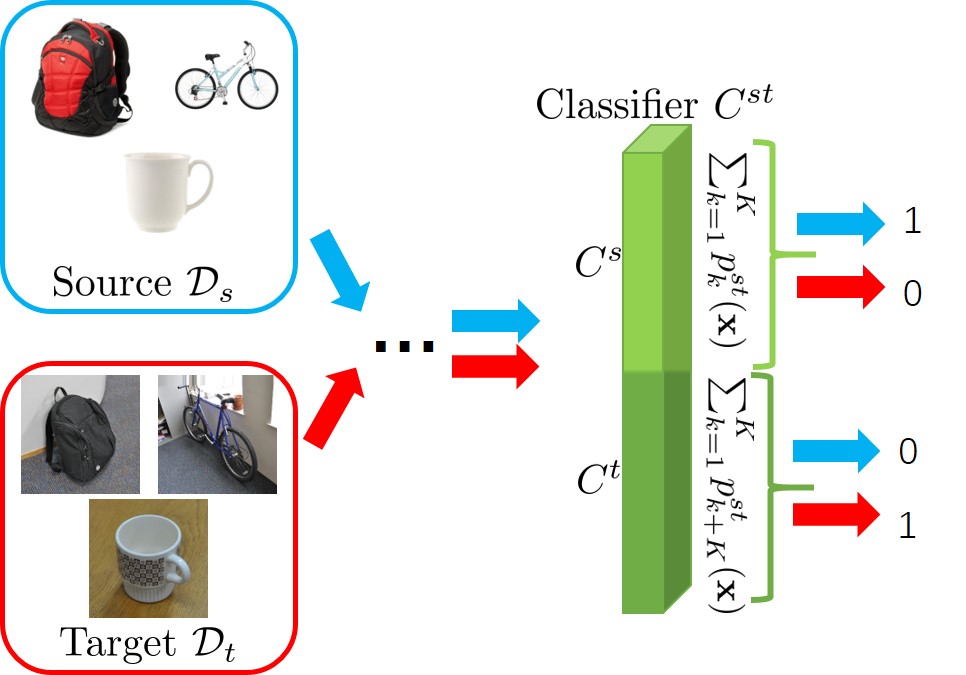}
\end{center}
   \caption{An intuitive presentation of how the loss (\ref{Loss_domain_stc}) differentiates the target classifier $\mathnormal{C}^t$ from the source classifier $\mathnormal{C}^s$.}
\label{fig:discriminator_illu}
\end{figure}

\vspace{0.01cm}
\noindent\textbf{Domain Discrimination}

Both $\mathnormal{C}^s$ and $\mathnormal{C}^t$ are trained using the labeled source samples. To differentiate between them, we leverage the constructed classifier $\mathnormal{C}^{st}$ in the SymNet. We train $\mathnormal{C}^{st}$ using the following two-way cross-entropy loss:
\begin{align} \label{Loss_domain_stc}
\noindent \notag \min_{\mathnormal{C^{st}}} {\cal{E}}_{domain}^{st} (G, C^{st}) = &- \frac{1}{n_t} \sum_{j=1}^{n_t} \mathrm{log}(\sum_{k=1}^{K} p^{st}_{k+K}(\mathbf{x}_j^t)) \\ & - \frac{1}{n_s} \sum_{i=1}^{n_s} \mathrm{log}(\sum_{k=1}^{K} p^{st}_k(\mathbf{x}_i^s)), 
\end{align}
where $\sum_{k=1}^{K} p^{st}_k(\mathbf{x})$ and $\sum_{k=1}^{K} p^{st}_{k+K}(\mathbf{x})$ can be viewed as the probabilities of classifying an input sample $\mathbf{x}$ as the source and target domains respectively. The objective of the loss (\ref{Loss_domain_stc}) is intuitively illustrated in Figure \ref{fig:discriminator_illu}.

Ideally, for the total $2K$ neurons of $\mathnormal{C}^{st}$, the combined effect of imposing losses (\ref{Loss_task_sc}), (\ref{Loss_task_tc}), and (\ref{Loss_domain_stc}) would be to make the set of first $K$ neurons discriminative among task categories, the set of last $K$ neurons discriminative among task categories, and to make the two sets distinguishable from each other. For example, for a source sample $\mathrm{\mathbf{x}}^s$ of the category $k$, both $\mathnormal{C}^s$ and $\mathnormal{C}^t$ tend to make accurate predictions, and for $\mathnormal{C}^{st}$, the probability of $p^{st}_{k}$ would be larger than $p^{st}_{k+K}$, due to the use of loss (\ref{Loss_domain_stc}). Similarly, for a target sample $\mathrm{\mathbf{x}}^t$ of the category $k$, both $\mathnormal{C}^s$ and $\mathnormal{C}^t$ tend to make accurate predictions, and for $\mathnormal{C}^{st}$, the probability of $p^{st}_{k+K}$ would be larger than $p^{st}_k$.

\subsection{A Two-level Domain Confusion Training of Domain-Symmetric Networks}
\label{Sec:two_level_confusion}

Similar to existing methods, we adopt the general strategy of adversarial training to learn an invariant feature extractor $\mathnormal{G}$ for the SymNet. More specifically, we propose a novel two-level domain confusion method that is based on a domain-level confusion loss and a category-level confusion loss. The proposed two-level losses aim to maximally `` confuse'' the two domains in order to align the joint distributions of feature and category across them.

To have a category-level confusion loss, we again rely on labeled source samples. For a source sample of category $k$, we identify its corresponding pair of the $k^{th}$ and $(k+K)^{th}$ neurons in $\mathnormal{C}^{st}$, and use a cross-entropy between predictions on this neuron pair and uniform distribution, which gives the following objective to learn the feature extractor $\mathnormal{G}$:
\begin{align} \label{Loss:cate_conf_s}
\noindent \notag\min_{\mathnormal{G}} {\cal{F}}_{category}^{st} (G, C^{st}) = & - \frac{1}{2n_s} \sum_{i=1}^{n_s} \mathrm{log}(p^{st}_{y_i^s+K}(\mathbf{x}_i^s)) \\ & - \frac{1}{2n_s} \sum_{i=1}^{n_s} \mathrm{log}(p^{st}_{y_i^s}(\mathbf{x}_i^s)).
\end{align}

To have a domain-level confusion loss, we use the unlabeled target samples, since label information is unnecessary for confusion at the domain level. For a target sample, we simply use a cross-entropy between aggregated predictions from the two half sets of neurons in $\mathnormal{C}^{st}$, and uniform distribution, which gives the following objective to learn the feature extractor $\mathnormal{G}$:
\begin{align} \label{Loss:do_conf_t}
\noindent \notag\min_{\mathnormal{G}} {\cal{F}}_{domain}^{st} (G, C^{st})=& - \frac{1}{2n_t} \sum_{j=1}^{n_t} \mathrm{log}(\sum_{k=1}^{K} p^{st}_{k+K}(\mathbf{x}_j^t)) \\ & - \frac{1}{2n_t} \sum_{j=1}^{n_t} \mathrm{log}(\sum_{k=1}^{K} p^{st}_k(\mathbf{x}_j^t)).
\end{align}
Note that one may opt for another domain-level confusion loss by using labeled source samples. We note that effect of such an additional loss may have been subsumed by the category-level confusion loss (\ref{Loss:cate_conf_s}), which uses labeled source samples.

\subsubsection{Entropy Minimization Principle} \label{Sec:emp}
Entropy minimization principle \cite{em} is adopted by some domain adaptation methods \cite{rtn, importance_weight, dirt_t} to enhance discrimination of learned models for target data. In this work, we adapt this principle to the symmetric structure of our proposed SymNet. We propose the following entropy minimization objective that enhances discrimination among task categories by summing over the probabilities at each pair of category-corresponding neurons in $\mathnormal{C}^{st}$:
\begin{equation} \label{Loss:em_t}
\noindent \min_{\mathnormal{G}} {\cal{M}}^{st} (G, C^{st}) = - \frac{1}{n_t} \sum_{j=1}^{n_t} \sum_{k=1}^{K} q^{st}_k(\mathbf{x}_j^t) \mathrm{log} (q^{st}_k(\mathbf{x}_j^t)),
\end{equation}
where $q^{st}_k(\mathbf{x}_j^t)$ = $p^{st}_k(\mathbf{x}_j^t) + p^{st}_{k+K}(\mathbf{x}_j^t$), $k \in \{1, \dots, K\}$.
As suggested by \cite{importance_weight}, instead of using (\ref{Loss:em_t}) for updating both the feature extractor $\mathnormal{G}$ and the classifier $\mathnormal{C}^{st}$, the entropy minimization loss is only used here to update $\mathnormal{G}$, in order to reduce the side effect that due to large domain shift, target samples may be stuck into wrong category predictions in the early stage of training, and are difficult to be corrected later on.

\subsection{The Overall Training Objective of Domain-Symmetric Networks}
\label{SecMainObj}

Combining the losses (\ref{Loss_task_sc}), (\ref{Loss_task_tc}), and (\ref{Loss_domain_stc}) for updating classifiers,  (\ref{Loss:cate_conf_s}) and (\ref{Loss:do_conf_t}) of category- and domain-level confusion for updating the feature extractor  $\mathnormal{G}$, and also the regularizer (\ref{Loss:em_t}), we have the following training objective for a SymNet:
\begin{align} \label{Loss:overall}
\notag & \min_{C^s, C^t, C^{st}} {\cal{E}}_{task}^s (G, C^s) + {\cal{E}}_{task}^t (G, C^t) + {\cal{E}}_{domain}^{st} (G, C^{st}) \\
& \min_{G} {\cal{F}}_{category}^{st} (G, C^{st}) + \lambda ({\cal{F}}_{domain}^{st} (G, C^{st}) + {\cal{M}}^{st} (G, C^{st})),
\end{align}
where $\lambda \in [0,1]$ is a trade-off parameter to suppress noisy signals of ${\cal{F}}_{domain}^{st} (G, C^{st})$ and ${\cal{M}}^{st} (G, C^{st})$ at early stages of training. The ${\cal{F}}_{category}^{st} (G, C^{st})$ is noise-free since it is based on the labeled source samples.

\section{Experiments}
We evaluate our SymNets on unsupervised domain adaptation tasks of three benchmark datasets and investigate the effects of the components in detail. The codes are available at \url{http://sites.scut.edu.cn/GPI/main.psp} 

\begin{table*}[htb]
\centering
\caption{Accuracy (\%) on the Office-31 dataset \cite{office_31}. All methods are based on models adapted from a 50-layer ResNet.}
\label{Tab:office31}
\begin{tabular}{lccccccc}
\hline
Methods                          & A $\to$ W     & D $\to$ W     & W $\to$ D         & A $\to$ D     & D $\to$ A      & W $\to$ A     & Avg  \\
\hline
ResNet-50 \cite{resnet}           & 68.4$\pm$0.2  & 96.7$\pm$0.1  & 99.3$\pm$0.1     & 68.9$\pm$0.2  & 62.5$\pm$0.3   & 60.7$\pm$0.3   & 76.1 \\
GFK \cite{gfk}                  & 72.8$\pm$0.0  & 95.0$\pm$0.0  & 98.2$\pm$0.0     & 74.5$\pm$0.0  & 63.4$\pm$0.0   & 61.0$\pm$0.0   & 77.5 \\
TCA \cite{tca}                  & 72.7$\pm$0.0  & 96.7$\pm$0.0  & 99.6$\pm$0.0     & 74.1$\pm$0.0  & 61.7$\pm$0.0   & 60.9$\pm$0.0   & 77.6 \\
DAN \cite{dan}                  & 80.5$\pm$0.4  & 97.1$\pm$0.2  & 99.6$\pm$0.1     & 78.6$\pm$0.2  & 63.6$\pm$0.3   & 62.8$\pm$0.2   & 80.4 \\
RTN \cite{rtn}                  & 84.5$\pm$0.2  & 96.8$\pm$0.1  & 99.4$\pm$0.1     & 77.5$\pm$0.3  & 66.2$\pm$0.2   & 64.8$\pm$0.3   & 81.6 \\
RevGrad \cite{reverse_grad}     & 82.0$\pm$0.4  & 96.9$\pm$0.2  & 99.1$\pm$0.1     & 79.7$\pm$0.4  & 68.2$\pm$0.4   & 67.4$\pm$0.5   & 82.2 \\
ADDA \cite{adda}                & 86.2$\pm$0.5  & 96.2$\pm$0.3  & 98.4$\pm$0.3     & 77.8$\pm$0.3  & 69.5$\pm$0.4   & 68.9$\pm$0.5   & 82.9 \\
JAN-A\cite{jan}                  & 86.0$\pm$0.4  & 96.7$\pm$0.3  & 99.7$\pm$0.1     & 85.1$\pm$0.4  & 69.2$\pm$0.3   & 70.7$\pm$0.5   & 84.6 \\      
MADA \cite{mada}                & 90.0$\pm$0.1  & 97.4$\pm$0.1  & 99.6$\pm$0.1     & 87.8$\pm$0.2  & 70.3$\pm$0.3   & 66.4$\pm$0.3   & 85.2 \\
iCAN \cite{collaborative}       & 92.5          & 98.8          & 100.0            & 90.1          & 72.1           & 69.9           & 87.2 \\
Kang \emph{et al.} \cite{attention_alignment} & 86.8$\pm$0.2  & \textbf{99.3}$\pm$0.1  & \textbf{100.0}$\pm$.0     & 88.8$\pm$0.4  & 74.3$\pm$0.2   & \textbf{73.9}$\pm$0.2   & 87.2 \\
CDAN+E \cite{cada}              & \textbf{94.1}$\pm$0.1  & 98.6$\pm$0.1  & \textbf{100.0}$\pm$.0     & 92.9$\pm$0.2  & 71.0$\pm$0.3   & 69.3$\pm$0.3   & 87.7 \\
\hline
SymNets                     & 90.8$\pm$0.1  & 98.8$\pm$0.3  & \textbf{100.0}$\pm$.0 & \textbf{93.9}$\pm$0.5 & \textbf{74.6}$\pm$0.6 & 72.5$\pm$0.5   & \textbf{88.4}  \\ 
\hline
\end{tabular}
\end{table*}

\begin{table*}[htb]
\centering
\caption{Accuracy (\%) on the ImageCLEF-DA dataset \cite{ImageCLEFDA}. All methods are based on models adapted from a 50-layer ResNet.}
\label{Tab:ImageCLEF}
\begin{tabular}{lccccccc}
\hline
Methods                         & I $\to$ P     & P $\to$ I     & I $\to$ C         & C $\to$ I    & C $\to$ P      & P $\to$ C      & Avg  \\
\hline
ResNet-50 \cite{resnet}           & 74.8$\pm$0.3  & 83.9$\pm$0.1  & 91.5$\pm$0.3     & 78.0$\pm$0.2  & 65.5$\pm$0.3   & 91.2$\pm$0.3   & 80.7 \\

DAN \cite{dan}                  & 74.5$\pm$0.4  & 82.2$\pm$0.2  & 92.8$\pm$0.2     & 86.3$\pm$0.4  & 69.2$\pm$0.4   & 89.8$\pm$0.4   & 82.5 \\

RevGrad \cite{reverse_grad}     & 75.0$\pm$0.6  & 86.0$\pm$0.3  & 96.2$\pm$0.4     & 87.0$\pm$0.5  & 74.3$\pm$0.5   & 91.5$\pm$0.6   & 85.0 \\ 

    
MADA \cite{mada}                & 75.0$\pm$0.3  & 87.9$\pm$0.2  & 96.0$\pm$0.3     & 88.8$\pm$0.3  & 75.2$\pm$0.2   & 92.2$\pm$0.3   & 85.8 \\
iCAN \cite{collaborative}       & 79.5          & 89.7          & 94.7             & 89.9          & 78.5           & 92.0           & 87.4 \\
CDAN+E \cite{cada}              & 77.7$\pm$0.3  & 90.7$\pm$0.2  & \textbf{97.7}$\pm$0.3     & 91.3$\pm$0.3  & 74.2$\pm$0.2   & 94.3$\pm$0.3   & 87.7 \\
\hline
SymNets    & \textbf{80.2}$\pm$0.3  & \textbf{93.6}$\pm$0.2 & 97.0$\pm$0.3 & \textbf{93.4}$\pm$0.3 & \textbf{78.7}$\pm$0.3  & \textbf{96.4}$\pm$0.1  &\textbf{89.9}  \\ 
\hline
\end{tabular}
\end{table*}

\begin{table*}[htb]
\begin{center}

\caption{Accuracy (\%) on the Office-Home dataset \cite{office_home}. All methods are based on models adapted from a 50-layer ResNet.}
\label{Tab:Office-Home}
\begin{tabular}{L{22.2mm}C{7.7mm}C{7.7mm}C{7.9mm}C{7.7mm}C{7.7mm}C{7.9mm}C{7.7mm}C{7.7mm}C{7.7mm}C{7.9mm}C{7.9mm}C{7.9mm}C{7.9mm}C{9mm}}
\hline
Methods                         &Ar$\to$Cl &Ar$\to$Pr &Ar$\to$Rw &Cl$\to$Ar &Cl$\to$Pr &Cl$\to$Rw &Pr$\to$Ar &Pr$\to$Cl &Pr$\to$Rw &Rw$\to$Ar &Rw$\to$Cl &Rw$\to$Pr    & Avg  \\
\hline
ResNet-50 \cite{resnet}           &34.9      & 50.0     & 58.0      & 37.4      & 41.9      & 46.2     & 38.5     & 31.2     & 60.4     & 53.9     & 41.2     & 59.9 &46.1 \\

DAN \cite{dan}                  & 43.6     & 57.0     & 67.9      & 45.8      & 56.5      & 60.4     & 44.0     & 43.6     & 67.7     & 63.1     & 51.5     &74.3  &56.3 \\

RevGrad \cite{reverse_grad}     & 45.6     & 59.3     & 70.1      & 47.0      & 58.5      & 60.9     & 46.1     & 43.7     & 68.5     & 63.2     &51.8      & 76.8 &57.6 \\ 

    
CDAN+E \cite{cada}             & \textbf{50.7}     & 70.6     & 76.0     & 57.6       & 70.0      & 70.0     & 57.4     & \textbf{50.9}     &77.3      &70.9      & \textbf{56.7}     &81.6 &65.8 \\
\hline
SymNets                     & 47.7 & \textbf{72.9} & \textbf{78.5}   & \textbf{64.2} &\textbf{71.3} &\textbf{74.2} &\textbf{64.2} &48.8 &\textbf{79.5} &\textbf{74.5} &52.6 &\textbf{82.7} & \textbf{67.6}  \\ 
\hline
\end{tabular}
\end{center}
\end{table*}

\subsection{Setup}

\noindent \textbf{Office-31} The office-31 dataset \cite{office_31} is a standard benchmark dataset for domain adaptation, which contains $4,110$ images of $31$ categories shared by three distinct domains: \textit{Amazon} (\textbf{A}), \textit{Webcam} (\textbf{W}) and \textit{DSLR} (\textbf{D}). We follow the common evaluation protocol on all six adaptation tasks. 

\noindent \textbf{ImageCLEF-DA} The ImageCLEF-DA dataset \cite{ImageCLEFDA} is a benchmark dataset for ImageCLEF 2014 domain adaptation challenge, which contains three domains: \textit{Caltech-256} (\textbf{C}), \textit{ImageNet ILSVRC 2012} (\textbf{I}) and \textit{Pascal VOC 2012} (\textbf{P}). For each domain, there are $12$ categories and $50$ images in each category. The three domains in this dataset are of the same size, which is a good complementation of the Office-31 dataset where different domains are of different sizes. We evaluate all methods on all six adaptation tasks. 

\noindent \textbf{Office-Home} The Office-Home dataset \cite{office_home} is a very challenging dataset for domain adaptation, which contains $15,500$ images from $65$ categories of everyday objects in the office and home scenes, shared by four significantly different domains: Artistic images (\textbf{Ar}), Clip Art (\textbf{Cl}), Product images (\textbf{Pr}) and Real-World images (\textbf{Rw}). We evaluate all methods on all $12$ adaptation tasks.

We compare our SymNets with shallow domain adaptation methods \cite{tca, gfk} and the state-of-the-art deep domain adaptation methods \cite{dan,rtn,reverse_grad,adda, jan, mada, SimNet, attention_alignment, cada}. We follow standard evaluation protocols for unsupervised domain adaptation \cite{reverse_grad, dan}: all labeled source samples and all unlabeled target samples are used for training. The average classification accuracy and the standard error of each adaptation task are reported on three random experiments. Our SymNets and all comparative methods are based on models adapted from a 50-layer ResNet \cite{resnet}. Especially, the deep representations output by the layer $pool$5 of ResNet are used as features for shallow methods.

We implement our SymNets based on PyTorch. A 50-layer ResNet pre-trained on the ImageNet dataset \cite{imagenet}, which excludes the last FC layer, is adopted as the feature extractor $\mathnormal{G}$. We fine-tune the feature extractor $\mathnormal{G}$ and train a classifier $\mathnormal{C}^{st}$ from scratch through back propagation. The learning rate of the classifier $\mathnormal{C}^{st}$ is $10$ times that of the feature extractor $\mathnormal{G}$. All parameters are updated by stochastic gradient descent (SGD) with momentum of $0.9$. The batch size is set to $128$. We follow \cite{reverse_grad} to employ the annealing strategy of learning rate and the progressive strategy of $\lambda$: the learning rate is adjusted by $\eta_p = \frac{\eta_0}{(1+\alpha p)^{\beta}}$, where $p$ is the progress of training epochs linearly changing from $0$ to $1$, $\eta_0 = 0.01$, $\alpha = 10$ and $\beta = 0.75$, which are optimized to promote convergence and low error on source samples; $\lambda$ is gradually changed from $0$ to $1$ by $\lambda_p = \frac{2}{1+\mathrm{exp}(-\gamma \cdot p)} - 1$, where $\gamma$ is set to $10$ in all experiments. Our classification results are obtained from the target task classifier $\mathnormal{C}_{t}$ unless otherwise specified, and the comparison between the performance of the source and target task classifiers is illustrated in Figure \ref{fig:convergence}.

\begin{table*}[htb]
	\centering
	\caption{Ablation experiments on the Office-31 dataset \cite{office_31}. All methods are based on models adapted from a 50-layer ResNet. Please refer to the main text for the detail definitions of these methods. 
	}
	\label{Tab:ablation_study}
	\begin{tabular}{lccccccc}
		\hline
		Methods                                       & A $\to$ W     & D $\to$ W     & W $\to$ D         & A $\to$ D    & D $\to$ A      & W $\to$ A     & Avg  \\
		\hline
		ResNet-50 \cite{resnet}                         & 79.9$\pm$0.3  & 96.8$\pm$0.4  & 99.5$\pm$0.1     & 84.1$\pm$0.4  & 64.5$\pm$0.3   & 66.4$\pm$0.4   & 81.9 \\
		ResNet-50 (Adding Em) \cite{resnet}            & 89.3$\pm$0.1  & \textbf{99.0}$\pm$0.1  & \textbf{100.0}$\pm$.0     & 89.2$\pm$0.7  & 73.4$\pm$0.1   & 69.0$\pm$0.2   & 86.6 \\
		Domain Confusion \cite{domain_confusion}       & 83.0$\pm$0.1  & 98.5$\pm$0.3  & 99.8$\pm$0.0     & 83.9$\pm$0.0   & 66.9$\pm$0.4   & 66.4$\pm$0.1   & 83.1 \\
		Domain Confusion (Adding Em) \cite{domain_confusion} & 89.8$\pm$0.7 & \textbf{99.0}$\pm$0.2  & \textbf{100.0}$\pm$.0    & 90.1$\pm$0.3   & 73.9$\pm$0.7  & 69.0$\pm$0.8   & 87.0 \\
		\hline
		SymNets (w/o ${\cal{E}}_{task}^{t}$)            & 75.3$\pm$0.9  & 95.9$\pm$0.2   & 99.6$\pm$0.2      & 75.1$\pm$0.9     & 60.2$\pm$0.3    & 62.7$\pm$0.7 & 78.1 \\
		SymNets (w/o ${\cal{M}}^{st}$)	                & 87.9$\pm$0.1  & 98.4$\pm$0.2  & 99.9$\pm$0.1     & 90.8$\pm$0.5  & 67.4$\pm$0.6  & 69.7$\pm$0.7 & 85.7 \\
		SymNets (w/o confusion)                         & 89.2$\pm$0.6  & \textbf{99.0}$\pm$0.3 & \textbf{100.0}$\pm$.0     & 93.8$\pm$0.3   & 73.7$\pm$0.2  & 65.9$\pm$0.6   & 86.9 \\
		SymNets (w/o category confusion)                & 89.9$\pm$0.6  & 98.1$\pm$0.1 & 99.8$\pm$0.0     & 93.7$\pm$0.5    & 71.9$\pm$0.2  & \textbf{73.5}$\pm$0.1   & 87.8 \\
		SymNets                                        & \textbf{90.8}$\pm$0.1  & 98.8$\pm$0.3  & \textbf{100.0}$\pm$.0 & \textbf{93.9}$\pm$0.5 & \textbf{74.6}$\pm$0.6 & 72.5$\pm$0.5   & \textbf{88.4}  \\ 
		\hline
	\end{tabular}
\end{table*}

\subsection{Results} \label{Sec:results}
The classification results on the Office-31 \cite{office_31}, ImageCLEF-DA \cite{ImageCLEFDA} and Office-Home \cite{office_home} datasets are reported in Table \ref{Tab:office31}, Table \ref{Tab:ImageCLEF} and Table \ref{Tab:Office-Home}, respectively. For fair comparison, results of other methods are either directly reported from their original papers if available or quoted from \cite{cada}. Our SymNets outperform all state-of-the-art methods on three benchmark datasets, highly affirming the effectiveness of our SymNets in aligning the joint distributions of feature and category across domains. It is compelling that our SymNets substantially enhance the classification accuracies on difficult adaptation tasks (e.g. \textbf{A} $\to$ \textbf{D} and \textbf{D} $\to$ \textbf{A}) and the challenging dataset (e.g. Office-Home). 
The Office-Home dataset is a very challenging dataset for domain adaptation due to following reasons as described in its original paper \cite{office_home}: (1) the number of categories is large in each domain; (2) different domains are visually very dissimilar; (3) the in-domain classification accuracy is low. 
Especially, the presence of large number of categories prejudices the domain alignment methods \cite{reverse_grad, dan, adda} for their ignorance of the alignment between corresponding categories of the two domains. It is desirable that our SymNets dramatically improve the performance on most adaptation tasks, demonstrating the efficiency of our proposed two-level domain confusion training of SymNets in aligning the joint distributions of feature and category across domains. 

\subsection{Analysis} \label{Sec:analysis}
\paragraph{Ablation Study}

\begin{figure*}
	\subfigure[ResNet-50]{
		\begin{minipage}[t]{0.24\linewidth}
			\centering
			\includegraphics[width=0.7\linewidth] {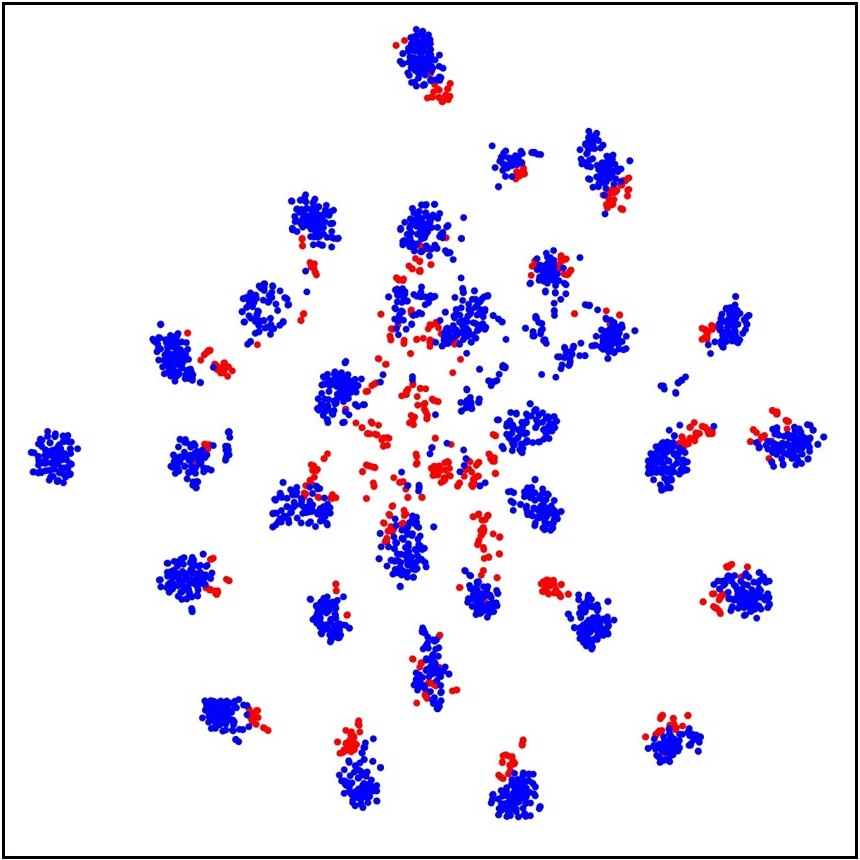}
			\label{Fig:sne_baseline}
	\end{minipage}}
	\hfill
	\subfigure[Domain Confusion]{
		\begin{minipage}[t]{0.24\linewidth}
			\centering
			\includegraphics[width=0.7\linewidth] {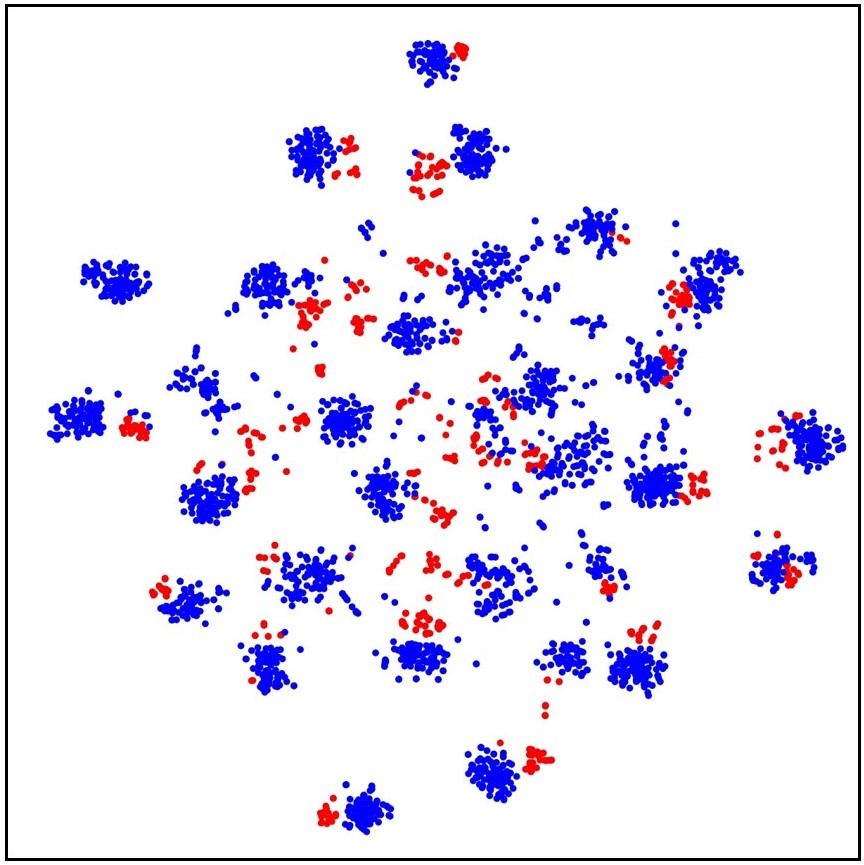}
			\label{Fig:sne_confusion}
	\end{minipage}}
	\hfill
	\subfigure[Domain Confusion (Adding Em)]{
		\begin{minipage}[t]{0.24\linewidth}
			\centering
			\includegraphics[width=0.7\linewidth] {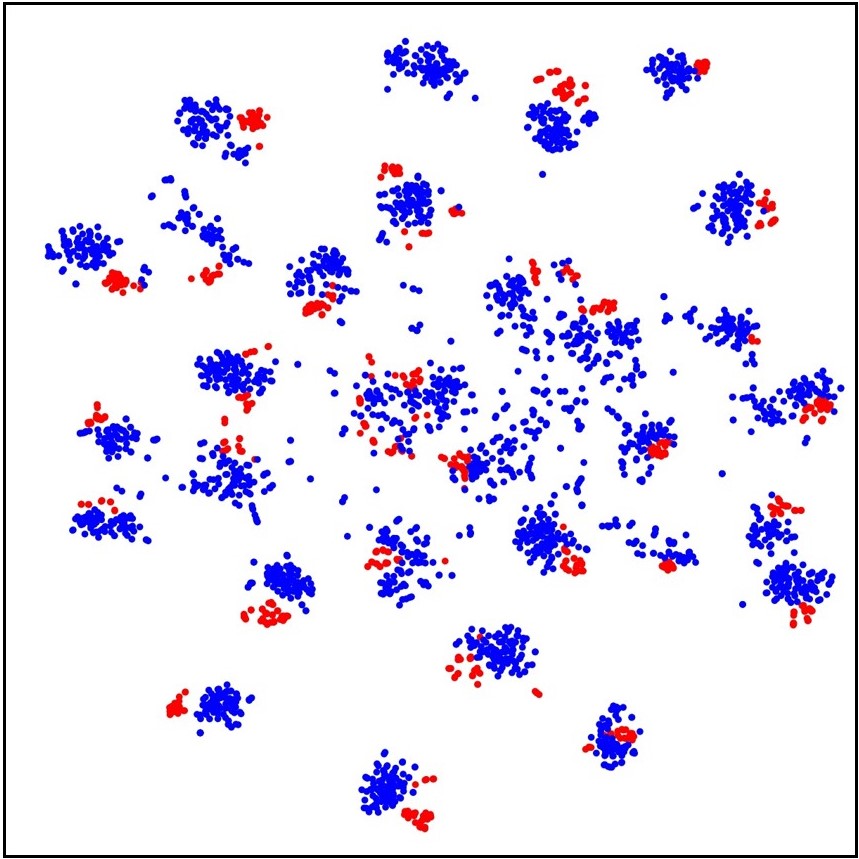}
			\label{Fig:sne_confusion}
	\end{minipage}}
	\hfill
	\subfigure[SymNets]{
		\begin{minipage}[t]{0.24\linewidth}
			\centering
			\includegraphics[width=0.7\linewidth] {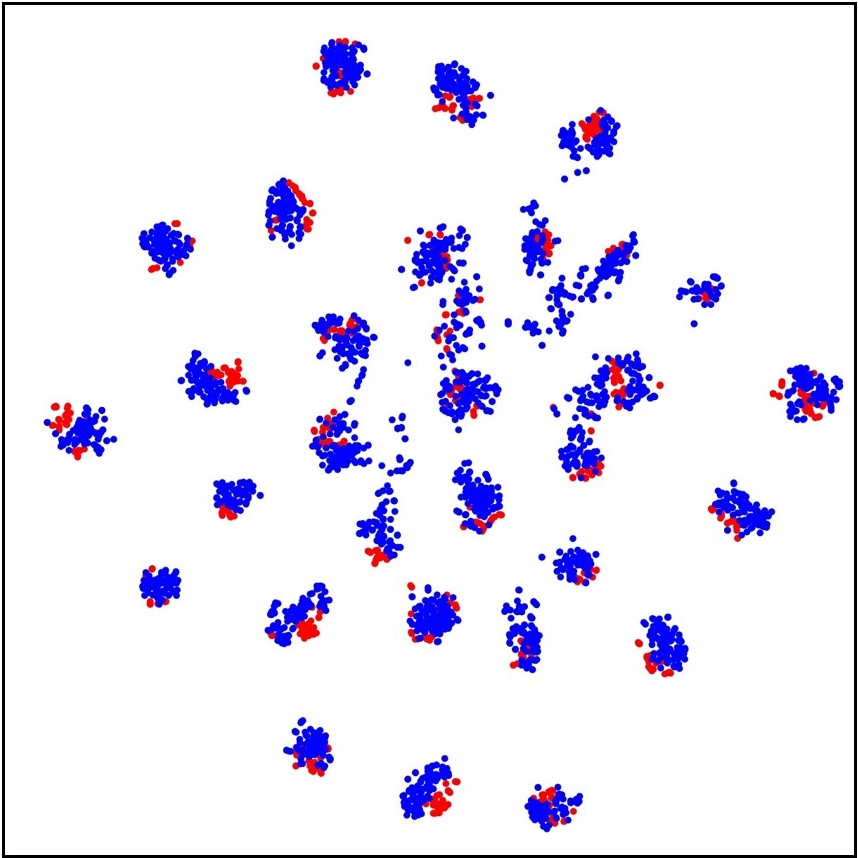}
			\label{Fig:sne_hada}
	\end{minipage}}
	
	\caption{The t-SNE visualization of feature representations learned by (a) ResNet-50, (b) Domain Confusion, (c) Domain Confusion (Adding Em) and (d) SymNets. Note that the blue and red points are samples from the source domain \textbf{A} and target domain \textbf{W} respectively.}
	\label{Fig:sne}
\end{figure*}

In this section, we conduct ablation experiments on the Office-31 dataset \cite{office_31} to investigate the effects of different components in our SymNets, which are based on models adapted from a 50-layer ResNet. We begin with the simplest baseline that fine-tunes on source samples the ResNet-50 model that is pre-trained on the ImageNet dataset \cite{imagenet}, which is denoted as ``ResNet-50''. To find out how the existing domain confusion method introduced in Section \ref{Sec:SymNet} performs, we conduct the experiment using the adversarial objective of domain confusion (\ref{Loss_confusion}), which is denoted as ``Domain Confusion''. To make it clear how our adopted entropy minimization loss presented in Section \ref{Sec:emp} can help the above two baselines, we additionally optimize the entropy minimization loss of target samples over their feature extractors and denote them as ``ResNet-50 (Adding Em)'' and ``Domain Confusion (Adding Em)'' respectively. To investigate how different components in our SymNets benefit the adaptation performance, we remove the cross-domain category supervised loss ${\cal{E}}_{task}^t (G, C^t)$ (\ref{Loss_task_tc}) and the entropy minimization loss ${\cal{M}}^{st} (G, C^{st})$ (\ref{Loss:em_t}) 
from the overall adversarial training objective (\ref{Loss:overall}), the training settings of which are denoted as ``SymNets (w/o ${\cal{E}}_{task}^t$)''  and ``SymNets (w/o ${\cal{M}}^{st}$)''
, respectively. Note that classification accuracies for SymNets (w/o ${\cal{E}}_{task}^t$) are obtained from the source task classifier $\mathnormal{C}^{s}$ due to the inexistence of the direct supervision signals in target task classifier $\mathnormal{C}^{t}$. Besides, to explore the effects of our proposed two-level domain confusion losses, we degenerate the category-level confusion loss ${\cal{F}}_{category}^{st}$ (\ref{Loss:cate_conf_s}) for source samples to a domain-level one: 
\begin{align}
\min_{\mathnormal{G}} & - \frac{1}{2n_s} \sum_{i=1}^{n_s} \mathrm{log}(\sum_{k=1}^{K} p^{st}_k(\mathbf{x}_i^s))  - \frac{1}{2n_s} \sum_{i=1}^{n_s} \mathrm{log}(\sum_{k=1}^{K} p^{st}_{k+K}(\mathbf{x}_i^s)), 
\end{align}
the training setting of which is denoted as ``SymNets (w/o category confusion)''. We remove the domain-level confusion loss ${\cal{F}}_{domain}^{st}$ (\ref{Loss:do_conf_t}) for target samples from the overall adversarial training objective (\ref{Loss:overall}) and degenerate the category-level confusion loss ${\cal{F}}_{category}^{st}$ (\ref{Loss:cate_conf_s}) for source samples to a general category classification loss:
\noindent\begin{align}
\noindent\min_{\mathnormal{G}} - \frac{1}{2n_s} \sum_{i=1}^{n_s} \mathrm{log}(p^{s}_{y_i^s}(\mathbf{x}_i^s)) - \frac{1}{2n_s} \sum_{i=1}^{n_s} \mathrm{log}(p^{t}_{y_i^s}(\mathbf{x}_i^s)), 
\end{align}
the training setting of which is denoted as ``SymNets (w/o confusion)''.

\begin{figure}
	\begin{center}
		\includegraphics[width=0.83\linewidth] {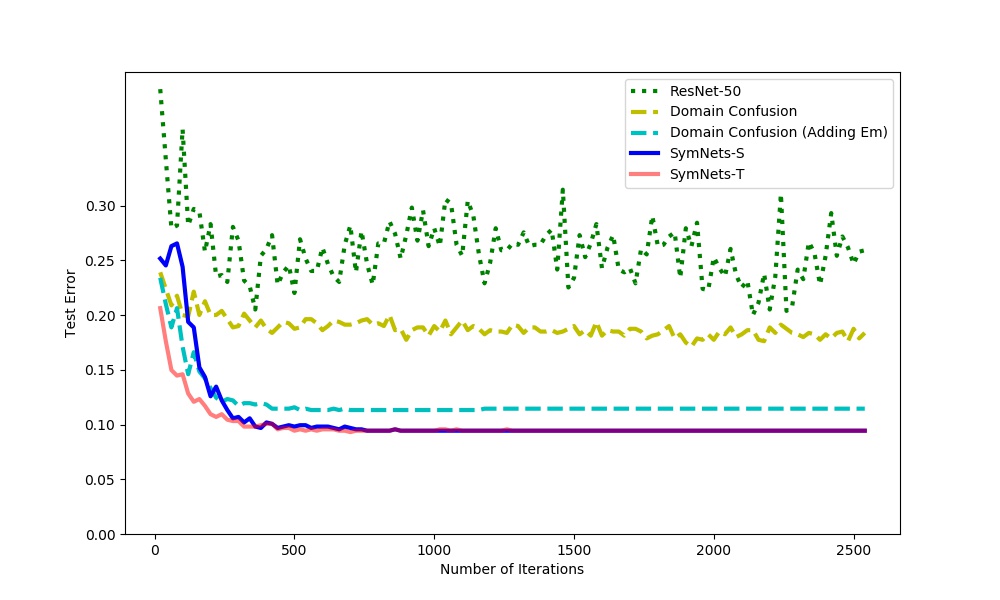}
	\end{center}
	\caption{Convergence performance on the adaptation task of \textbf{A} $\to$ \textbf{W} by ResNet-50, Domain Confusion, Domain Confusion (Adding Em), and the source and target task classifiers of our SymNets, which are denoted as SymNets-S and SymNets-T respectively.}
	\label{fig:convergence}
\end{figure}

The results are reported in Table \ref{Tab:ablation_study}. ``Domain Confusion'' performs much better than ``ResNet-50'', and ``SymNets (w/o category confusion)'' improves over ``SymNets (w/o confusion)'', testifying the effectiveness of the domain-level confusion in the feature alignment. Observed that the performance of ``SymNets (w/o ${\cal{E}}_{task}^t$)'' suffers a slump of $10.3\%$, manifesting the importance of the cross-domain category supervised loss ${\cal{E}}_{task}^t (G, C^t)$ (\ref{Loss_task_tc}) to learn a well-performed target task classifier in the two-level confusion training of SymNets. SymNets enhances the adaptation performance over ``SymNets (w/o category confusion)'', certifying the usefulness of our proposed category-level confusion in the alignment between corresponding categories of the two domains. 
The entropy minimization loss ${\cal{M}}^{st} (G, C^{st})$ (\ref{Loss:em_t}) consistently improves both the two baselines of ``ResNet-50'' and ``Domain Confusion'' and our SymNets by a large margin in performance, demonstrating its efficacy. By fair comparison, our SymNets achieve the best result among these ablation experiments, confirming their excellent effect in aligning the joint distributions of feature and category across domains.

\paragraph{Convergence Performance}
We compare the convergence performances of task classifiers of $\mathnormal{C}^{s}$ and $\mathnormal{C}^{t}$ in our SymNets 
with ``ResNet-50'', ``Domain Confusion'' and ``Domain Confusion (Adding Em)'' in Figure \ref{fig:convergence}. The test errors of different methods on the adaptation task \textbf{A} $\to$ \textbf{W} are reported. We observe that our SymNets converge much smoother. The convergence performances of the source task classifier $\mathnormal{C}^{s}$ and target task classifier $\mathnormal{C}^{t}$ are expected. At the beginning of adversarial training, the performance of the target task classifier on target samples is better, since the source task classifier and target task classifier are specified to corresponding domains. As the training proceeds, the joint distributions of feature and category are gradually aligned across domains, thus the performance of two task classifiers almost converge to the same level. 

\paragraph{Feature Visualization}
We visualize the network activations from feature extractors of ``ResNet-50'', ``Domain Confusion'', ``Domain Confusion (Adding Em)'' and our SymNets on the adaptation task \textbf{A} $\to$ \textbf{W} by t-SNE \cite{sne} in Figure \ref{Fig:sne}. The source and target domains are not well aligned for features of ``ResNet-50''. For features of ``Domain Confusion'', the two domains are better aligned, however, the data structure of target samples is scattered and the shared categories across domains are not well aligned. For features of ``Domain Confusion (Adding Em)'', the data structure of target samples is well preserved, but the shared categories across domains are not well aligned. For features of our SymNets, the shared categories across domains are perfectly aligned while different categories are well distinguished. The effectiveness of the two-level domain confusion training of SymNets in aligning joint distributions of feature and category across domains is verified intuitively.


\section{Conclusion}
We propose a novel adversarial learning method termed domain-symmetric networks (SymNets) to overcome the limitation in aligning the joint distributions of feature and category across domains via two-level domain confusion losses. The category-level confusion loss improves over the domain-level one by driving the learning of intermediate network features to be invariant at the corresponding categories of the two domains. As a component of the SymNets, an explicit target task classifier is learned through a cross-domain training scheme. Experiments on three benchmark datasets verify the efficacy of our proposed SymNets.


\vspace{0.2cm}
\noindent \small \textbf{Acknowledgment.} This work is supported in part by the National Natural Science Foundation of China (Grant No.: 61771201), and the Program for Guangdong Introducing Innovative and Enterpreneurial Teams (Grant No.: 2017ZT07X183).

{\small
\bibliographystyle{ieee_fullname}
\bibliography{egbib}
}

\end{document}